# A Survey on Distributed Evolutionary Computation

Wei-Neng Chen, *Senior Member*, *IEEE*, Feng-Feng Wei, Tian-Fang Zhao, Kay Chen Tan, *Fellow*, *IEEE*, and Jun Zhang, *Fellow*, *IEEE*

*Abstract*—The rapid development of parallel and distributed computing paradigms has brought about great revolution in computing. Thanks to the intrinsic parallelism of evolutionary computation (EC), it is natural to implement EC on parallel and distributed computing systems. On the one hand, the computing power provided by parallel computing systems can significantly improve the efficiency and scalability of EC. On the other hand, data are collected and processed in a distributed manner, which brings a novel development direction and new challenges to EC. In this paper, we intend to give a systematic review on distributed EC (DEC). First, a new taxonomy for DEC is proposed from top design mechanism to bottom implementation mechanism. Based on this taxonomy, existing studies on DEC are reviewed in terms of purpose, parallel structure of the algorithm, parallel model for implementation, and the implementation environment. Second, we clarify two major purposes of DEC, i.e., improving efficiency through parallel processing for centralized optimization and cooperating distributed individuals/sub-populations with partial information to perform distributed optimization. Third, noting that the latter purpose of DEC is an emerging and attractive trend for EC with the booming of spatially distributed paradigms, this paper gives a systematic definition of the distributed optimization and classifies it into dimension distributed-, data distributed-, and objective distributed-optimization problems. Formal formulations for these problems are provided and various DEC studies on these problems are reviewed. We also discuss challenges and potential research directions, aiming to enlighten the design of DEC and pave the way for future developments.

## I. INTRODUCTION

THE past decade has witnessed the rapid development and widespread applications of parallel and distributed computing systems such as supercomputing, cloud computing, and edge computing. By aggregating massive distributed computing resources, these computing systems provide powerful computing capability for handling complex and large-scale computation applications. In addition, aided by the Internet of Things (IoTs) technology, things are interconnected and thus a huge amount of data is distributedly collected, stored and needed to be processed, leading to the era of Big Data. Consequently, the increasing amount of data and the distributed mechanism of information processing pose an urgent need and new challenges to the development of intelligent algorithms on these novel distributed computing systems.

Evolutionary computation (EC), as an active research field in computational intelligence, has forged ahead in the last few decades and has been widely used in solving complex optimization problems [1]. EC is characterized by simulating biological evolution or intelligent behavior of social animals to evolve a population of solutions. It mainly contain evolutionary algorithms like genetic algorithm (GA) [2], genetic programming (GP) [3], and swarm intelligence algorithms like ant colony optimization (ACO) [4], particle swarm optimization (PSO) [5], etc. Although EC is powerful to solve complicated optimization problems with regular or irregular search space due to its good exploration, exploitation, self-learning, self-organizing, adaptability, etc. [6], [7], [8], iterative evolution paradigm reduces computing efficiency and therefore hinders its applications in solving large-scale or expensive optimization problems. Fortunately, EC algorithms have intrinsic parallelism, which provide a great opportunity to combine EC with novel parallel and distributed computing systems.

With the rapid development of high-performance computing, such as multiprocessor systems [9], graphical processing unit (GPU) [10], clusters [11], etc., a considerable amount of research effort has been devoted to parallel and distributed EC. They mainly aim to improve efficiency of EC for complex centralized optimization problems, in which information can be fused in a center [12], [13]. The most widely used method is to implement the algorithm in dimension level distributed parallelism or population level distributed parallelism [14]. For

Manuscript received XXX; revised XXX; accepted XXX. This work was supported in part by the National Key Research and Development Project, Ministry of Science and Technology, China, under Grant 2018AAA0101300; in part by the National Natural Science Foundation of China under Grants 61976093 and 61873097; in part by the National Research Foundation of Korea under Grant NRF-2021H1D3A2A01082705; and in part by the Guangdong Natural Science Foundation Research Team under Grant 2018B030312003.

Wei-Neng Chen and Feng-Feng Wei are with the School of Computer Science and Engineering, South China University of Technology, Guangzhou 510006, China, and with the Guangdong-Hong Kong Joint Innovative Platform of Big Data and Computational Intelligence, South China University of Technology, Guangzhou, 510006, China. (e-mail: cschenwn@scut.edu.cn, fengfeng_scut@163.com).

Tian-Fang Zhao is with the School of Journalism and Communication, Jinan University, Guangzhou, 510006, China (e-mail: tianfang09@foxmail.com).

Kay Chen Tan is with the Department of Computing, The Hong Kong Polytechnic University, Hong Kong SAR (e-mail: kctan@polyu.edu.hk).

Jun Zhang is with the Department of Electrical and Electronic Engineering, Hanyang University, Ansan 15588, South Korea (e-mail: junzhang@ieee.org).

example, Zhan et al. [15] implemented EC algorithms on GPU to realize fined-grained dimension distributed parallelism. Tan et al. [16] designed a coevolutionary algorithm in population level distributed parallelism on networked computers. Jia et al. [17] proposed a double-layer distributed algorithm in both dimension level distributed parallelism and population level distributed parallelism on multiprocessor systems. They can greatly improve efficiency without sacrificing the performance of algorithms. Besides, parallel and distributed EC has also been applied in industry, such as electromagnetics [18], virtual network embedding [19], etc.

With the advent of Big Data era and development of IoTs, novel distributed computing techniques such as edge computing [20], cloud computing [21], multi-agent systems (MASs) [22], etc. are bursting and have attracted much research interests. Consequently, many information is generated, collected, processed, stored in a distributed manner and cannot be fused due to high computational cost, limited storage space and information privacy. Therefore, distributed optimization problems are emerging. Different from centralized optimization, individuals in EC for distributed optimization has no global perspective for the whole problem and needs to cooperate. Compared to parallel and distributed EC for centralized optimization, though there are fewer studies on distributed EC (DEC) for distributed optimization, it is becoming an attractive research direction nowadays.

In conclusion, on the one hand, the development of distributed computing paradigms makes a breakthrough in computer science. The combination of EC and parallel computing platforms has significantly improved efficiency and scalability of EC for large-scale optimization problems. On the other hand, applications of novel distributed computing systems and MASs give birth of many distributed optimization problems, and DEC has preliminarily shown potential in solving these complex distributed optimization problems. However, there is still a lack of systematic review of related work. Therefore, this paper gives a comprehensive survey of DEC. Main contributions of this paper are as follows.

Firstly, this paper reviews DEC from two perspectives, a) DEC for centralization optimization and b) DEC for distributed optimization. DEC for centralization optimization assumes that individuals or subpopulations of the algorithm share the same memory or data, and the goal is usually to take advantage of parallel and distributed computing platforms to improve optimization efficiency. On the contrary, DEC for distributed optimization assumes that individuals or subpopulations of the algorithm has its own memory to save local data and information. They cannot share memory and need to cooperate through specific cooperative strategies to achieve global optimization. Due to the difference in problem characteristics, design for these two kinds of problems is greatly different.

Secondly, this paper proposes a new taxonomy to review DEC, which is from top design mechanism to bottom implementation mechanism. Specifically, this paper reviews DEC algorithms in terms of purpose and problem, parallel structure of the algorithm, parallel model for implementation, and implementation environment. Purposes and implementation environments are different due to that these two kinds of problems have characteristics of centralized and distributed respectively. Parallel structure and model are similar for these two kinds of problems. Parallel structure is the core difference between DEC algorithms and traditional serialized EC algorithms. It refers to the modules and structures in the algorithm that can be executed in parallel. Parallel model refers to the interconnection network model about how computing resources are organized and communicated in the process of algorithm implementation in parallel.

Thirdly, this paper gives a systematic definition of distributed optimization problems and introduces potential research trends of DEC. Most existing DEC algorithms are proposed for centralized optimization. They are confined in problems with global objective function and centralized decision variables. However, distributed optimization arises with the booming of IoTs, in which data are distributedly collected, stored and processed. There is no global perspective for the whole problem. Specifically, we classify distributed problems as dimension distributed problems, data distributed problems, and objective distributed problems. For each kind of problems, this paper gives a detailed introduction and mathematical formulation, and surveys the pioneer DEC studies.

The rest of this paper is organized as follows. Section II elaborates the taxonomy of this paper. Section III to section V

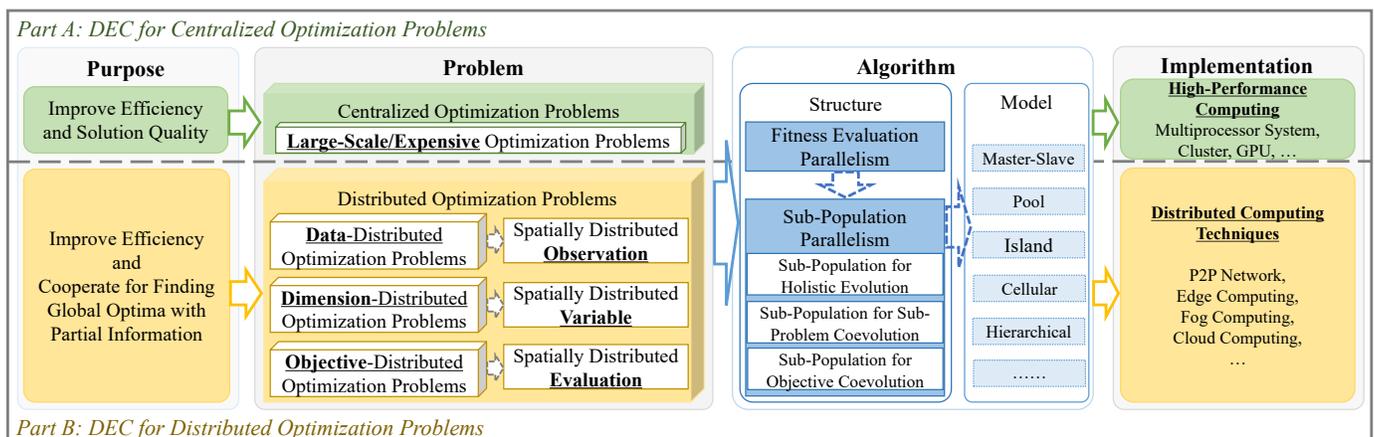

**Fig. 1.** The taxonomy of this paper, which includes problem and purpose, parallel structure and model, and implementation environment.

review parallel structures of the algorithm, parallel models for implementation and implementation environments respectively. Section VI gives a detailed introduction to emerging distributed optimization problems and reviews existing DEC for them. Section VII discusses challenges and potential research directions of DEC. Section VIII summarizes the whole paper.

## II. TAXONOMY

The new taxonomy to review DEC is depicted in Fig. 1. According to problem properties, we classify problems solved by DEC into two categories: centralized optimization problems in upper green areas and distributed optimization problems in lower yellow areas. For different kinds of problems, purposes, problem properties and implementation environments are different. However, categories of parallel structures in DEC algorithms and parallel models for implementation are similar, which are shown in blue parts of Fig. 1. According to problems and purposes, this paper gives an extensive review from top algorithm design to bottom implementation environment and introduces DEC for distributed optimization in detail.

**Problem and Purpose**. In general, there are two different purposes for DEC.

The first purpose is to take advantage of the computing power provided by parallel and distributed computing systems to improve the computing efficiency of EC. When facing large-scale and expensive optimization problems, EC requires a lot of time to maintain the iterative evolution process of the population and is therefore very time-consuming. In this case, parallel and distributed computing can make the evolution of the population or the evaluation of the objective function executed in parallel, and thus can save execution time.

The second purpose is to apply EC to solve distributed optimization problems in distributed systems. Traditional optimization problems are usually centralized, that is, all individuals in the population share the same dataset and focus on the same goal to solve a global optimization problem. However, with the development of IoT and distributed computing, distributed optimization problems have emerged. Different from centralized optimization problems, data are collected by different individuals or agents in distributed optimization problems, and the data cannot be aggregated due to transmission efficiency or privacy. Thus, the individuals have to collaborate to achieve global optimization. According to which part of the problem definition is given in a distributed manner, distributed optimization problems can be categorized into three classes: dimension distributed optimization problems, data distributed optimization problems, and objective distributed optimization problems. As the name implies, they have spatially distributed variables, observations, and evaluations respectively.

**Parallel Structure of the Algorithm**. Parallel structure refers the modules or structures of an algorithm that can be executed simultaneously. For DEC, parallel structures can be generally classified into two categories, fitness evaluation parallelism and sub-population parallelism.

Fitness evaluation parallelism is the naivest parallelism, in which the fitness evaluations of each individual are executed in parallel. Sub-population parallelism is that the evolution operators of each sub-population or even each individual are performed in parallel. According to the function of sub-populations, we can further divide sub-population parallelism into sub-population parallelism for holistic evolution, sub-population parallelism for sub-problem coevolution, and sub-population parallelism for objective coevolution. As the name suggesting, sub-population parallelism for holistic evolution is that multiple sub-populations parallelly explore the whole search space. Sub-population parallelism for sub-problem coevolution is that the original problem is decomposed into several sub-problems with parts of decision variables and multiple sub-populations cooperatively optimize these sub-problems for global optima. Sub-population parallelism for objective coevolution is specially designed for problems with multiple objectives. Different sub-populations focus on different objectives and they need to cooperate with each other to find Pareto front.

TABLE I
AN INVESTIGATION OF DEC STUDIES

| Purpose | Problem | Number of Population | Parallelism Structure | Algorithm Model | Implementation Environment | Representative Papers |
|---|---|---|---|---|---|---|
| Improve Efficiency and Solution Quality | Centralized Optimization Problems | Single | Fitness Evaluation Parallelism | Island Model | GPU | [169], etc. |
| | | Multiple | Sub-Population Parallelism for Holistic Evolution | | | [109], [129], [170], etc. |
| | | | | Cellular Model | Multiprocessor System | [79], [82], [83], etc. |
| | | | | Master-Slave Model | Cluster | [106], etc. |
| | | | Sub-Population Parallelism for Cooperative Coevolution | Island Model | | [105], [140], etc. |
| Cooperate for Finding Global Optima with Partial Information | Distributed Optimization Problems | Single | Sub-Population Parallelism for Objective Coevolution | Master-Slave Model | Multi-Agent System | [153], etc. |
| | | Multiple | Sub-Population Parallelism for Holistic Evolution | | Cloud Computing | [131], [143], etc. |
| | | | | Cellular Model | Multi-Agent System | [90], [91], [92], etc. |
| | | | | Island Model | P2P Network | [139], etc. |

**Parallel Model for Implementation**. Parallel model refers to the interconnection network model about how computing resources are organized and communicated for algorithm implementation in parallel. Generally, parallel models for implementation can be divided into two categories, coarse-grained models and fine-grained models. Coarse-grained models have larger subcomponents in each processor for computation such as subpopulation whereas fine-grained models have smaller subcomponents such as individuals or even dimensionalities. These processors are connected in a specific network. Characteristics of the network, such as full connected or partial connected, cyclic or acyclic, directed or undirected, etc. determine the organization and communication of processors. Accordingly, there are four widely used types of parallel models, master-salve model, island model, cellular model, and hierarchical model. Master-slave model and cellular model are fine-grained models and island model is a fine-grained model. Hierarchical model is a combination of more than one above mentioned models and therefore can be a kind of combined coarse-grained and fine-grained model.

**Implementation Environment**. Implementation environment is the basis for developing DEC algorithms. In the literature, parallel and distributed EC have been implemented on both high-performance computing environments such as GPU computing systems [10], multiprocessor systems [9], clusters [11], etc., and spatially distributed computing systems such as P2P networks [23], edge computing [20], fog computing [24], cloud computing [21], etc.

In high-performance computing environments, memories and data are shared and all processors focus on the same goal to solve a global optimization problem. It is very common to use high-performance computing environments to improve efficiency. In spatially distributed computing environments, data are collected, stored, processed by distributed processors and it is hard to aggregate data due to transmission efficiency or privacy. Processors need to cooperatively search for global optima. It is of vital importance to design effective and efficient communication strategies for coevolution.

A comprehensive investigation of DEC from purpose to implementation environment is given in Table I, which is helpful to facilitate the understanding of new taxonomy and development of DEC.

## III. PARALLEL STRUCTURE OF THE ALGORITHM

For EC, the evaluation of the objective functions and the evolution of each subpopulation (or individual) are parallelizable. Different modules for parallelism lead to different parallel structures. Therefore, we classify the parallel structure of DEC into fitness evaluation parallelism and sub-population parallelism to give a review.

### A. Fitness Evaluation Parallelism

Fitness evaluation parallelism is to evaluate fitness through multiple processors. There are two ways to implement fitness evaluation parallelism. The first one is pure fitness evaluation parallelism, in which evolution is conducted on a center and evaluations are conducted through multiple processors. The other one is that fitness evaluation parallelism is cooperated with sub-population parallelism. It is a common parallelism structure since conducting evaluations with sub-populations is more efficient.

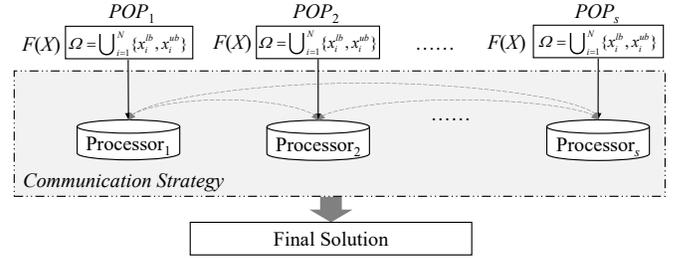

**Fig. 2.** The illustration of sub-population parallelism for holistic evolution.

In fitness evaluation parallelism, a biggest problem is fitness evaluation budget allocation. Fitness evaluation budget allocation is to determine how many fitness evaluations are allocated in each processor. Apart from equal allocation strategy, in which processors have the same budget of fitness evaluation resources, some adaptive fitness evaluation allocation strategies have been proposed. Ren *et al*. [25] developed a computation resource allocation strategy to reasonably assign fitness evaluations to sub-populations on the basis of their contributions to fitness improvement. Jia *et al*. [17] designed a periodic contribution calculating method and allocated computing resources based on calculated contribution. Li *et al*. [26] proposed an adaptive resource allocation among distributed sub-populations. It aims to adaptively allocate more fitness evaluations on well-performing parts. Theoretical analyses are provided to show effectiveness of the proposed method.

### B. Sub-Population Parallelism

#### 1) Sub-Population Parallelism for Holistic Evolution

Let denote the original minimization problem as follows:
$$\min F(X), \ X \in \Omega, \quad \Omega = \bigcup_{i=1}^{N} \{x_i^{lb}, x_i^{ub}\} \quad (1)$$

where $F$ is the objective function to be optimized. $X$ is a $D$ dimensional decision variable vector in large search space $\Omega$, which is constrained by the upper bound $x_i^{ub}$ and lower bound $x_i^{lb}$ of each dimension. Sub-population parallelism for holistic evolution is to divide the whole population into multiple sub-populations and each sub-population evolves in the whole search space. As shown in Fig. 2, there are $s$ sub-populations and each sub-population is to optimize problem $F(X)$ in the whole se arch space $\Omega$. Specifically, when each population contains one individual, it is the individual parallelism. The main research point in sub-population parallelism for holistic evolution is migration strategy.

Migration strategy is when and what to exchange and update sub-populations. It greatly influences the performance since cooperation plays an importance role. Generally, migration has synchronous manner and asynchronous manner. Synchronous migration is to wait for all sub-populations receiving information whereas asynchronous migration is to receive information whenever it arrives in a sub-population.

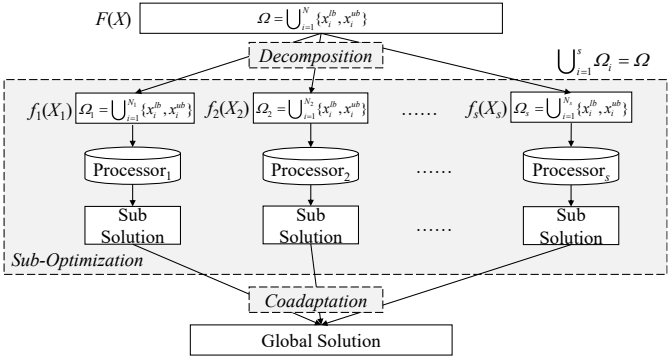

**Fig. 3.** The illustration of sub-population parallelism for sub-problem coevolution.

Synchronous migration is a widely used manner since it is effective and easy to implement. Alba *et al.* [12], [27] implemented EAs in a parallel way, in which migration synchronously happens. De Falco *et al.* [28] proposed an invasion-based migration model for distribute DE that sub-populations consecutively send and receive individuals in a synchronous manner.

However, it is a waste of time to wait for arrival of information in each sub-population. Therefore, asynchronous migration is arising and becomes more popular in sub-population parallelism. Apolloni *et al.* [29] proposed an asynchronous migration, in which individuals to be migrated are randomly chosen by a selection function. De Falco *et al.* [27] proposed an adaptively asynchronous migration mechanism. They use a probabilistic-based method to determine whether migration is triggered. Once migration is happened, the best individual from source sub-population replaces the worst individual in target sub-population. Besides, De Falco *et al.* [30] exploited diversity of asynchronous migration in distributed DE. Jakobović *et al.* [31] designed an asynchronous elimination global parallel EA and demonstrated it has higher efficiency compared to other parallel algorithms. Izzo *et al.* proposed an asynchronous island-model and tested on POSIX Threads application programming interface. Zhan *et al.* [32] developed an adaptive migration strategy, which can deliver robust solution with better performance among sub-populations. Fernández *et al.* [33] analyzed advantages and disadvantages of both synchronous and asynchronous migration based on distributed GP. They found that asynchronous manner is more suitable for multiprocessor systems. Alba and Troya [34] investigated synchronous and asynchronous migration among sub-populations. They found that asynchronous migration outperformed synchronous migration in terms of time.

*2) Sub-Population Parallelism for Sub-Problem Coevolution*

Sub-population parallelism for sub-problem coevolution is usually to solve large-scale problems, which are extremely hard to find global optimum due to the curse of dimensionality. To reduce the optimization difficulty, the original problem is divided into several sub-problems, which are easier to optimize. Divide and conquer is a commonly used technique to split the large-scale problem into smaller scale problems [35]. The integration of divide-and-conquer strategy and EC algorithms gives the birth of cooperative coevolution (CC), which is the

**Algorithm 1** *Pseudo Codes of DG*

**Input**: Ungrouped variable set $\Omega = \bigcup_{i=1}^{N}\{x_i^{lb}, x_i^{ub}\}$

    **repeat**
1:     $\Omega_s = x_{1st}$ in $\Omega$, $\Omega = \Omega - x_{1st}$;
2:     **if** $\Omega$ is not empty
3:         **for** $j=2:|\Omega|$
4:             **if** *interaction*($x_{1st}$, $x_j$) is **true**
5:                 $\Omega_s = \Omega_s \cup x_j$, $\Omega = \Omega - x_j$;
6:             **end if**;
7:         **end for**;
8:     **end if**;
    **until** $\Omega$ is not empty

**Output**: The grouped variable sets $\Omega_1 = \bigcup_{i=1}^{N_1}\{x_i^{lb}, x_i^{ub}\}$, $\Omega_2 = \bigcup_{i=1}^{N_2}\{x_i^{lb}, x_i^{ub}\},\ldots,\Omega_s = \bigcup_{i=1}^{N_s}\{x_i^{lb}, x_i^{ub}\}$ most widely used method for sub-problem coevolution. As shown in Fig. 3, there are *s* sub-populations and each sub-population is to optimize one sub-problem $f_i(X_i)$ in sub search space $\Omega_i$ with several dimensionalities. The union of all sub search spaces cover the whole search space. Sub-population parallelism for sub-problem coevolution contains decomposition, sub-optimization and coadaptation, in which decomposition and coadaptation are two important steps.

1) Decomposition

Decomposition is to decompose the large-scale decision variable $X$ into $s$ smaller scale decision variables $X_1, X_2, \ldots, X_s$. Correspondingly, the original search space $\Omega$ is decomposed into sub search spaces $\Omega_1, \Omega_2, \ldots, \Omega_s$. The union of all sub search spaces $\bigcup_{i=1}^{s}\Omega_i = \Omega$ is the whole search space $\Omega$. Each smaller scale decision variable is regarded in a sub-problem $f_i(X_i)$. If all variables are independent from each other, the decomposition strategy has little influence on optimization results. This kind of problem is known as *separable* problem and easier to optimize. If variables have some relationship, the decomposition strategy has great influence to optimization results. This kind of problem is known as *non-separable* problem, which are common in practical applications. It has been demonstrated that near-optimal decomposition is possible to quantify the contribution of a sub-problem to global fitness [36].

Generally, decomposition strategies can be classified into static decomposition strategies and dynamic decomposition strategies. Static decomposition strategies are conducted before evolution and the decomposition is kept fixed during optimization. The representative static decomposition strategies contain random grouping [37], fast interdependency identification (FII) [38], differential grouping (DG) [39] *etc.* Random grouping is to fix the number of groups and group size [37]. Variables are randomly grouped in a group. Random grouping has lower time complexity and good performance in separable optimization problems. However, it cannot handle non-separable optimization problems well. FII divides detection into two stages. Firstly, separable and non-separable variables are identified. Then, the relationship of non-separable variables is further detected. DG proposed by Omidvar *et al.* [39] is a most widely used grouping method for non-separable optimization problems. Pseudo codes of DG are shown in Algorithm 1. Firstly, the first variable in ungrouped variable set is regarded as a new group and excluded from the set. Then, interaction between the first variable and all other variables in ungrouped variable set are examined. If the examined interaction is true, the detected variable is grouped with the first variable and excluded from the remaining ungrouped variables. If no variable is interacted with the first variable, the first variable itself is a group. This process is repeated until ungrouped variable set is empty. After the proposal of DG, many variants have been developed, such as XDG [40], GDG [41], DG2 [42], RDG [43], RDG2 [44], DDG [45], MDG [46], etc.

Dynamic decomposition strategies can dynamically make a decomposition during optimization. The representation dynamic decomposition strategies contain frequent random grouping [47], delta grouping [48], dynamic grouping (DyG) [49], knowledge-based dynamic variable decomposition (KDVD) [50], *etc*. Omidvar *et al*. [47] proposed a more frequent random grouping to increase the frequency of random grouping whereas the budget of fitness evaluation does not increase. Further, they considered to break the limit on improvement interval of interacting variables and proposed delta grouping [48]. Zhang *et al*. [49] proposed a dynamic grouping method (DyG) to adaptively allocation computational resources to elitist groups. Cai *et al*. [50] focused on bilevel multi-objective optimization problems and designed a knowledge-based dynamic variable decomposition (KDVD).

2) Coadaptation

Coadaptation between sub-problems is to share evolved information during fitness evaluation. How to select representative information from other sub-problems to form a complete individual for evaluation is of vital importance. Generally, coadaptation can be divided into best collaborator coadaptation, worst collaborator coadaptation, random collaborator coadaptation, elite collaborator coadaptation and strategy-based collaborator coadaptation. Let use denote *j*th individual in the *i*th sub-problem $\vec{x}_{i,j}$ to be evaluated as follows.

$$F(X) | X = \{\vec{x}_{1,*}, \vec{x}_{2,*}, ..., \vec{x}_{i,j}, ..., \vec{x}_{s,*}\} \quad (2)$$

where $\vec{x}_{o,*}$ ($o=1,2,...,s$ and $o \neq i$) are coadaptation information from other sub-problems. Best collaborator coadaptation means that $\vec{x}_{o,*}$ is the best individual in other sub-problems. It is one of the most widely used strategy in sub-problem parallelism [37], [39], [51]. Specifically, best collaborator coadaptation is quite suitable for separable problems, whereas it easily leads to local optima for non-separable problems [52]. On the contrary, worst collaborator coadaptation means that $\vec{x}_{o,*}$ is the worst individual in other sub-problems. Although worst collaborator coadaptation can provide more alternative candidates, it is not an optimal one and not widely used in CC [53]. Random collaborator coadaptation means that $\vec{x}_{o,*}$ is a randomly selected individual from other sub-problems [53]. It can significantly improve diversity of the population and prevent premature convergence [16]. However, high randomness may reduce convergence speed [54]. Elite collaborator coadaptation means that $\vec{x}_{o,*}$ is randomly chosen from elite pool, which is composed of the best *K* individuals from each sub-problem [55].

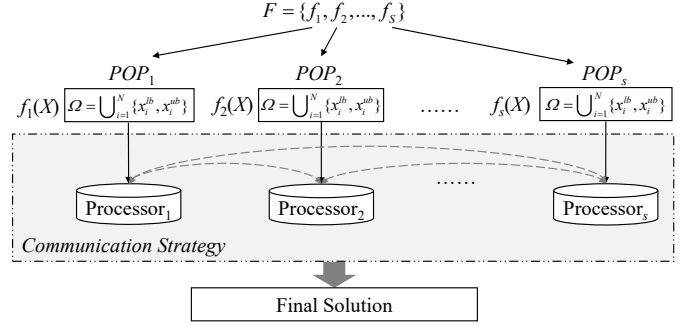

Fig. 4. The illustration of sub-population parallelism for objective coevolution.

Specifically, if *K* is 1, elite collaborator coadaptation is the same as the best collaborator coadaptation. Elite collaborator coadaptation can prevent premature convergence due to the diversity of elite pool [16]. Strategy-based collaborator coadaptation means that $\vec{x}_{o,*}$ is selected through some specific strategy from each sub-problem, such as roulette selection [56] or tournament selection [57]. Strategies give more chooses for candidates and different strategies have different advantages.

3) *Sub-Population Parallelism for Objective Coevolution*

Sub-population parallelism for objective coevolution is to use different sub-populations to evolve different objectives. As shown in Fig. 4, the optimization problem has *s* objectives and each objective is optimized by one sub-population in all dimensionalities $f_i(X)$. Sub-population parallelism for objective coevolution has not gained as much attention as formal two sub-population parallelisms out of two reasons. Firstly, DEC usually studies problems in which fitness can be accessed at the same time. In other words, all objective values can be acquired by all agents. Secondly, there should be more than one objective for sub-population parallelism to coevolution, which intensifies study challenges. Some efforts have been done for optimization with heterogeneous objective evaluations, in which evaluations of objectives or constraints differ in computational time [58]. Allmendinger *et al*. [59] divided the population into fast population and slow population to acquire objectives with short latencies and long latencies respectively. Lewis *et al*. [60] studied master-slave model using multi-objective PSO for heterogeneous optimization. They extended the proposed method to distributed environment. Scott and De Jong [61] proposed an asynchronous EA based on master-slave model to alleviate evaluation-time bias. Yagoubi *et al*. [62] developed an asynchronous evolutionary multi-objective algorithms with heterogeneous evaluation costs.

IV. PARALLEL MODEL FOR IMPLEMENTATION

In the literature, four types of parallel models are commonly used for DEC, including master-slave models, island models, cellular models, and hierarchical models shown in Fig. 5. The bigger circles or ellipses are distributed components to conduct computation. Black dots are individuals and arrows mean communications between distributed components.

*A. Master-Slave Model*

Master-slave model is kind of fine-grained model. As shown in Fig. 5 a), master-slave model is composed of one master

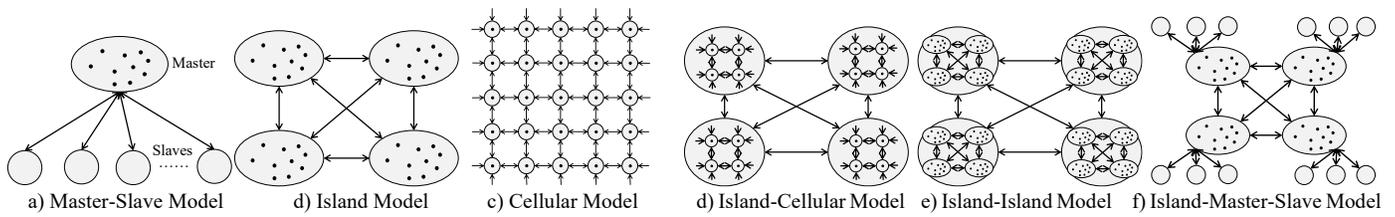

a) Master-Slave Model  d) Island Model  c) Cellular Model  d) Island-Cellular Model  e) Island-Island Model  f) Island-Master-Slave Model

**Fig. 5.** The illustration of algorithm models. Black dots represent individuals and arrows mean communications.

processor and several slave processors. Originally master-slave model is proposed for fitness evaluation parallelism, in which the master takes charge of evolution and sends individuals to slaves for the most time-wasting part in optimization, fitness evaluation [63]. Communication in master-slave models only occur between the slave processor and master processor. There is no direct communication among slave processors.

*1) Communication Strategies*

With the development of master-slave model variants, slaves are also in charge of parts of evolution tasks [64], [65]. To improve efficiency of master-slave model, asynchronous strategies are applied. Yang *et al*. [66] proposed an asynchronous adaptive communication method based on request-response mechanism. Communication between the master processor and slave processors is adaptively triggered during evolution. Said and Nakamura [67] proposed an asynchronous communicate strategy that slave processors can start a new iteration as soon as the terminal condition is satisfied. If slave processors have completed the phase before the master processor completes the evaluation required for the next phase, they need to be re-initialized according to local optimal solution with the completed phase.

*2) Improved Master-Slave Models*

Besides, efforts have been made to improve effectiveness of master-slave distributed EAs. Ismail [68] applied master-slave to GA and modified the model to improve efficiency. In the modified model, slave processors are responsible for not only fitness evaluation process but also mutation process and parts of crossover process. The master processor only takes charge of selection and distribution of the population. Yu and Zhang [69] proposed a hierarchical GA based on master-slave model. Global search is carried on the master processor whereas local search is executed on slave processors. Chromosomes in the master layer indicate the search center of slave layers and they evolve based on search results of slave layers. Lorion *et al*. [70] proposed an agent based parallel PSO which is composed of a coordination agent and several swarm agents. The swarm agents are in calculation state or trade state. With the help of niching method, each swarm agent sells particles far away from center point and buys particles close to center point.

*3) Pool Model*

Pool model is a quite special kind of master-slave model in which slaves conduct evolution rather than the master. The master acts as a pool to save the evolution information of slaves. Slaves can asynchronously evolve by reading global information from the pool and storing local information to the pool. Roy *et al*. [71] proposed a special kind of master-slave model, called pool model, and applied it in GA. In this model, the master processor is only responsible for storing individuals in the pool. Slave processors extract individuals from the pool for calculation and evolution. After that, the updated individuals are rewritten into the pool. Yang *et al*. [66] proposed a distributed PSO based on the pool model. The master processor maintains the global best individual, and slave processors carry evolution. An elite-guided learning strategy is designed to update particles. Elite particles in the current swarm and the best historical solutions found by different slave processors are employed to guide the update of particles in each slave processor.

*B. Island Model*

The structure of island model is shown in Fig. 5 b). As a coarse-grained parallel model, island model decomposes the whole population to several sub-populations and distributes each sub-population to a computational node. Therefore, island model is generally used for sub-population parallelism.

*1) Communication Strategies*

To improve efficiency of algorithms, asynchronous communication strategies are proposed to avoid unnecessary waiting. Du *et al*. [72] proposed an asynchronous distributed EDA for gene expression programming. In the algorithm, each processor evolves independently and communicates with other processors after several generations. A certain number of worse individuals in each processor are replaced by better individuals received from other processors. Zhan and Zhang [73] proposed an adaptive asynchronous migration strategy for PSO. The migration of swarm is executed independently according to search environment and current evolution state.

*2) Improved Island Models*

Island model was first introduced by Tanese *et al*. [74] in a distributed GA (DGA). Experimental results verified that the proposed DGA outperforms canonical serial GA in random Walsh polynomials optimization. Then, Belding [75] further validated the effectiveness of DGA on Royal Road class. After that, Romero and Cotta [76] applied island model to PSO and tested the influence of different migration parameters. The choices of migrating particles include random selecting and choosing the best particle. The replacement strategy includes always replace and replace if solution is better. The topology of island includes ring topology and complete topology. Apolloni *et al*. [29] applied unidirectional island model to DE in which migrating particles are random selected for replacement if they are better. Ishimizu and Tagawa [77] further explored the performance of island DE with different network topologies including ring, torus, hypercube, hierarchical. Zhang *et al*. [78] proposed a distributed DE combining Lamarckian learning and Baldwinian learning with Hooke–Jeeves local search strategy. In this model, sub-populations are connected by von Neumann topology, which is the same as topology structure of cellular model.

*C. Cellular Model*

Cellular models are a kind of fine-grained model, which distributes each individual in population to a computing node. As shown in Fig. 5 c), each individual is considered as a cell living in a lattice and can only communicate with its neighbors. Due to its easy implementation on parallel hardware, cellular model is popular [12], [79]. As neighbor lattices are overlapped, individuals with good performance can be spread its advantages to the whole population. Compared with coarse-grained island model, more processors are required in cellular model when the population size is large.

*1) Communication Strategies*

In early studies of cellular EA, cells communicate with each other and updated themselves synchronously. This mechanism guarantees algorithm stability, but sometimes leads to inefficiency. A kind of asynchronous cellular EC algorithm was proposed to solve this problem, which includes four kinds of asynchronous updating mechanisms: fixed directional sweep, fixed random sweep, random new sweep, and uniform choice [80]. All cells are updated by a predefined and fixed order in fixed directional sweep, by a random and fixed order in fixed random sweep, and by a random but unfixed order in random new sweep. In uniform choice, a random cell is chosen to update each time. Akat and Gazi [81] proposed an asynchronous cellular PSO in which cells can communicate with each other and be updated at different time instants. Outdated information is employed for updating particles. Each particle can directly communicate with its neighbors which change dynamically with time. When iteration of neighbor particle has not been finished, the current optimal solution is applied as a candidate of global best solution.

*2) Improved Cellular Model*

Modifications have been made to the initial cellular model to improve effectiveness of EC algorithms. Alba and Troya [82] defined a ratio to measure the relationship between topology and neighborhood in cellular EA, and proposed a cellular EA that can adjust the ratio adaptively. Nebro *et al*. [83] proposed a new cellular genetic algorithm called MOcell to deal with multi-objective optimization problems. Giacobini *et al*. [84] paid attention to the selection intensity in different kinds of cellular EA including asynchronous cellular EA, linear cellular EA [85], and toroidal cellular EA [86]. After that, they summarized the research about selection intensity of different kinds of cellular EA and employed the probability difference equation based on synchronous and asynchronous cell update strategy to derive the selection model [87].

As a fine-grained model, cellular model needs a large number of computational nodes to finish the distributed deployment of population. Nakashima *et al*. [88] proposed a coarse-grain cellular model, called combined cellular GA to settle this problem. In combined cellular GAs, several individuals are considered as cells live together in one lattice. Only cells in edges of the lattice can communicate with cells in edges of the neighbor lattice. [89].

*3) Multi-Agent Model*

Multi-agent model is a variant of cellular model [90], which has the same structure with cellular model, namely, distributing individuals to a lattice. Each lattice is considered as an agent and cooperates or competes with neighbors. Different from traditional cellular model, agents in multi-agent model have their own behaviors. The main behavioral strategies in multi-agent model include neighborhood competition or cooperation operator, evolution operator (determined by specific EA), and self-learning operator. Zhong *et al*. [90] combined multi-agent system with GA and proposed a multiagent GA (MAGA). Followed by MAGA, Zhao *et al*. [91] proposed a novel PSO based on multiagent system (MAPSO) to solve reactive power dispatch problem. After that, MAEA-CSPs was proposed to solve constraint satisfaction problems [92], and MAEA-CmOPs was proposed to solve combinatorial optimization problems [93]. In terms of communication strategy, Laredo *et al*. [94] proposed an asynchronous communication method for multi-agent model, which can adaptively set the migration rate according to network conditions. Ding *et al*. [95] designed a co-evolutionary quantum PSO based on multi-agent system, which aim to explore the search space and global best region during attribute reduction for big datasets. Zhang *et al*. [96] proposed an evolutionary multi-agent framework for equipment layout multi-objective optimization. Ji *et al*. [97] proposed a multi-agent evolutionary model to discover communities in a complex network.

*D. Hierarchical Model*

Hierarchical model is a combination of two or more models to take their advantages for both fitness evaluation parallelism and sub-population parallelism. For example, Fig. 5 d), e), f) give examples of island-cellular hierarchical model, island-island hierarchical model, and island-master-slave hierarchical model respectively. Folino *et al*. combined island and cellular models to study distributed GP algorithm for P2P computing [98] and pattern classification [99]. Herrera *et al*. [100] developed a distributed GA in island-island hierarchical model. Burczynski *et al*. [101], [102] and Lim *et al*. [103] proposed to combine island model with master-slave model. They decomposed population into several sub-populations, which are sent to slaves by the master. The upper layer is an island model for further cooperation.

## V. Implementation Environment

The implementation environment is the basis of developing DEC algorithms. In the literature, parallel and distributed EC algorithms have been implemented on both high-performance computing environment such as graphics processing units (GPU) computing systems, multiprocessor systems, clusters, etc., and spatially distributed computing systems such as P2P networks, cloud computing, etc.

*A. High-Performance Computing*

*1) Multiprocessor System and Cluster*

The multiprocessor system is a collection of more than one central processing unit (CPU) in one system [9] whereas cluster collects more than one CPU in two or more independent systems. They are popular implementation environments for DEC algorithms to improve efficiency. Vavpotic *et al*. [104] have described a multiprocessor approach to speed up the execution of EC algorithms with multiprocessor systems. They divide the population into several sub-populations and each sub-population evolves in a processor. It has been applied on Tomita automata 3 that speedup is proportional to the number of processors. Yang *et al.* [66] developed a distributed swarm optimizer on multiprocessor system. They realized nearly linear speedup as the number of processors increases in 1,000 up to 3,000-dimensional optimization problems. Zhan *et al.* [15] developed matrix-based GA and PSO in dimension level on CPU. For problems with 1,000 dimensionalities, matrix-based GA and PSO can reach 50 times speedup with 64 cores. Bazlur Rashid *et al.* [105] given a comprehensive review of DEC on MapReduce cluster to solve large-scale optimization problems within less computing time. Du *et al.* [106] proposed a parallel gene expression programming based on MapReduce cluster for solving symbolic regression. They parallelly executed operations such as fitness evaluation, crossover, mutation and selection and the speedup can reach 15 times with 64 processors.

*2) GPU*

GPU can execute lots of threads in parallel. Due to its powerful computational ability, GPU has been widely used to handle complex tasks with large computation burden [10].

Haghofer *et al.* [107] implemented the evolution strategy as island model on GPU for cell detection in microscopy images. The optimized parameters are set from 200 to 5,000 with maximal 6,000 pixels in experiments and higher test quality consumes longer GPU execution time. Laguna-SaNchez *et al.* [108] presented a parallel implementation of DE on NVIDIA GPU. They executed each individual in a thread and conduct experiments on 30-dimensional problems with different numbers of individuals. It has been demonstrated that speedup is much improved when individuals increase to 1024 with 15.000 iterations. Huang *et al.* [109] focused on large-scale evolutionary multitasks and proposed a GPU-based paradigm to simultaneously optimize hundreds of thousands of optimization requests. Specifically, they achieved the highest 1,426 times speedup for 1,000 tasks.

B. *Distributed Computing*

*1) P2P Network*

P2P network is a distributed infrastructure, which allows more than peers to connect and communicate with each other [23]. Due to its high adaptability, reliability and performance, P2P network has been involved in distributed optimization. Wickramasinghe *et al.* [110] designed an adaptively autonomous selection for EAs in P2P network. Results on N Queens problems, where the maximal number of N is 96, showed its viability and scalability. Laredo *et al.* [111][112] proposed a spatially structured EA consists of a population of evolvable agents (EvAg) by P2P protocol. They found EvAg has better scalability than sequential GA, generational GA and steady-state GA due to the distributed structure. Besides, EvAg requires a small amount of computational resources and has higher execution efficiency in solving problems with 2 to 4 traps. Besides, Laredo *et al.* [113] have investigated the influence of population size on execution time in P2P EAs. They conducted experiments in PeerSim Simulator, which is a P2P simulator, and found that P2P EAs have better scalability and higher efficiency. There are two conclusions, a) the population size scales with polynomial order with respect to the dimensionality and b) the execution time scales with fractional power with respect to the population size.

*2) Cloud Computing*

Cloud computing is a novel distributed computing mode, in which the computing takes place in data centers and processed via cloud [21]. With improved agility and time-to-value, higher scalability and cost-effectiveness, it is attracting more research interests. Malmir *et al.* [114] trained a multi-layer perceptron network using PSO for data mining on cloud computing. They found that parallel processing is not speed up the processing for small amount of data, but it good when facing with large amount of data and can classify data in a shorter elapsed time. Yu *et al.* [115] proposed a modified immune EA for medical data clustering and feature extraction under cloud computing environment. Experiments on Brest Cancer Wisconsin (Original) Data Set in UCI Machine Learning Repository showed that the proposed method can extract important attributes for clinical diagnosis and improve diagnosis speed and accuracy. Besides, with more than 10,000,000 samples to be clustered, the proposed method can achieve 85% accuracy.

C. *Applications of EC for Computing Modes*

Providing the implementation environments for EC is a combination of computing modes and EC. Conversely, EC is also used to optimize some problems in computing modes, such as the optimization of workflow scheduling, resource allocation, task offloading, energy consumption, and so on.

For high-performance computing, Pillai *et al.* [116] used a GA-based method to optimize scheduling of multiprocessor systems to handle the energy consumption. To solve the problem of load balance among processors, Bi *et al.* [117] proposed a distributed evolutionary game algorithm for multiprocessors selection. Handzel *et al.* [118], [119] used EC techniques to optimize task scheduling and allocation in heterogeneous multiprocessor system. Kim *et al.* [120] designed a GA for task allocation for multiprocessors. What is more, Zhou *et al.* [121] have given a theoretical study of EC algorithm for multiprocessor scheduling and allocation problem.

For spatially distributed computing, Anandaraj *et al.* [122] proposed an efficient and novel GA to solve network code minimization problem in P2P networks. Zuo and Zhang [123] used an evolutionary game model to improve efficiency of self-interested routing and overcome confusion condition caused by selfish routing. Dai *et al.* [124] developed a distributed algorithm based on evolutionary game to increase the usage of mobile cloud computing. Ye *et al.* [125] proposed an improved knee point driven EA to optimize workflow scheduling in cloud computing. Zhu *et al.* [126] developed a novel evolutionary multi-objective optimization scheme for work flow scheduling in cloud computing. Teijeiro *et al.* [127] developed an island-based parallel version of DE on Microsoft Azure public cloud platform.

D. *Discussions*

Implementing DEC on high-performance computing environments and spatially distributed computing environments both can greatly improve efficiency of DEC algorithms for timely cost problems. However, they also have many differences.

In high-performance computing environments, processors share the same global memory and they can frequently communicate with each other for information exchange [128]. DEC mainly aims to accelerate execution in parallel through massive computing resources. With global information shared among all processors, it is easy for processors to achieve consistence and convergence within less time. It is good at processing centralized optimization problems [14]. However, high-performance computing environments are generally big and costly. Processors are put in the same area and connected through the local network. It is costly for development, maintenance and extension. The execution time for solving problems is limited by the scale of computing resources.

When designing DEC algorithms in high-performance computing environments, researchers pay more attention on which modules and how to parallelly execute. The mostly used evaluation metric is the speedup ratio $R_s$ [13][63][129], which is calculated as follows:

$$R_s = \frac{T_1}{T_p} \quad (3)$$

where $T_1$ is the sequential execution time by 1 processor and $T_p$ is the parallel execution time by $p$ processors. It reflects the how many times the parallel execution can be faster than sequential execution as $p$ increases. Ideally, when $R_s=p$, in which $T_1= T_p \times p$, it is said that the algorithm makes full use of computing resources and reaches linear speedup ratio. In this circumstance, $T_p$ can reduce the corresponding times than $T_1$ with $p$ increases. However, it is hard to realize linear speedup ratio in most cases. To measure the utilization of computing resources, the efficiency $E$ is introduced, which is derived through $R_s$ and calculated as $E=R_s/p$. The algorithm with high $E$ represents that it has high computing resource utilization. Obviously, when the algorithm has linear speedup ratio, $E=100\%$. With the increasing of problem scale and the expansion of computing resources, scalability of the algorithm is also a concerned metric. If the computing resources and problem scale are expanded in the same proportion, and $E$ is not decreased, it can be regarded that the algorithm has good scalability on large-scale problems and large-scale parallel computing platforms.

In spatially distributed computing environments, processors only have their local memories and it takes high cost to communicate with each other [130][131]. Therefore, besides improving efficiency through massive computing resources, one important purpose of DEC is to cooperate distributed processors to search for global optima. It is appropriate to implement DEC on spatially distributed computing systems for distributed optimization problems. In spatially distributed computing environments, processors are located in diverse areas and connected through the Internet. Users generally just need to rent processors without maintenance, which has lower expense and higher scalability. However, with local information which are not disclosed to other processors, it is important to study the consistence and convergence among all processors within limited communications [132].

When designing DEC algorithms in spatially distributed computing environments, except for improving efficiency, researchers also concentrate on distributed communication [132][133]. Due to the high communication cost, researchers need to consider what, when and how to communicate. Communication volume and frequency are limited by transmission bandwidth. Frequent communication or large volume transmission can reduce efficiency and on the contrary, little communication between individuals or subpopulations may greatly influence the cooperation for global optimization. Moreover, data protection also should be taken into account, which is a growing need nowadays [134].

## VI. DEC for Distributed Optimization Problems

The above discussed DEC techniques are mainly used to solve centralized optimization problems with global objective functions and centralized decision variables. They are proposed to improve execution efficiency and solution quality through the cooperation of multiple processors. However, as the booming of Internet and big data, mass distributed optimization problems emerge, in which there is no global perspective for the whole problem. Main purpose of DEC for distributed optimization is to cooperatively search for global optimum, which heavily rely on information exchange between agents and multiple local decision makers. Typical applications include distributed storage of cloud data, IoTs with local sensing ability, edge computing, etc. These new problems and applications pose challenges to existing DEC techniques. Therefore, this section firstly gives an introduction to distributed optimization problems and then elaborates the definition and taxonomy of this kind of problems. Importantly, some pilot studies of DEC techniques are reviewed to pave the way for distributed optimization problems.

*A. Distributed Optimization Problems*

*1）Introduction*

Distributed optimization problems are an emerging kind of problems in intelligent optimization. Different from centralized optimization problems, which are processed in a centralized way, distributed optimization problems have natural characteristics of distribution. Firstly, data are collected in a distributed manner and it is costly expensive or even impossible to integrate all data in a fusion center. Secondly, computing is carried out in spatial distributed techniques. Distributed optimization problems refer to that lots of agents $a$ located in different places and are connected in a specific topology. They generate or collect data and process data independently. Fewer or even none of them own global information of the whole problem. Each agent only has its local information, which is incomplete. The purpose of agents is to optimize problems based on their own information and the received information. Data collection and computation of different agents are asynchronous and independent from each other. Only agents which are connected can conduct communications. Due to expensive communication cost and privacy protection, communications cannot be conducted frequently. The high spatial distribution and limited local information make it difficult for traditional optimization problems, such as simulated annealing [135], Tabu search [136], artificial neural networks (ANNs) [137], etc. to find global optimum. Therefore, distributed optimization problems propose great challenges of spatially local information and asynchronous processing to most existing optimization methods.

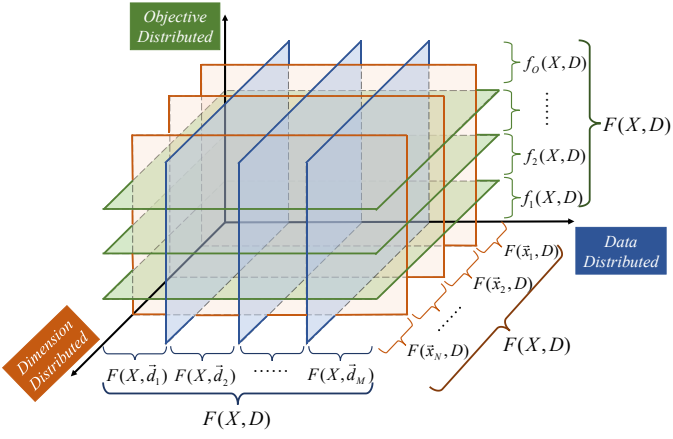

**Fig. 6.** The distributed information of distributed optimization problems. Blue planes divide the data into multiple distributed parts. Orange planes divide the dimension into multiple distributed parts. Green planes divide the objective into multiple distributed parts.

*2) Definition and Taxonomy*

Without loss of generality, a minimization optimization problem can be formulated as follows:

$$\min F(X,D), X \in \Omega \quad (4)$$

in which $F$ is objective that agents aim to cooperatively minimize and $D$ is environmental data set. $X$ is variable vector to be optimized and $\Omega$ is variable space of $X$. In distributed optimization, spatially distributed information can be environmental data $D$, variable dimensionality $X$ and objective $F$. Each distributed agent only has one part of information and the union of all agents forms the whole information.

As shown in Fig. 6, planes divide the whole information into different parts, which are owned by different agents. Specially, blue planes mean the division of data whereas spaces of dimension and objective are the whole. This is the first basic kind of problem: data distributed optimization problems and the formulation is as follows:

$$\min F(X, \vec{d}_i), X \in \Omega, i = 1, 2, ..., M$$
$$D = \{\vec{d}_1, \vec{d}_2, ..., \vec{d}_i, ..., \vec{d}_M\} \quad (5)$$

Orange planes in Fig. 6 mean the division of dimension whereas spaces of data and objective are the whole. This is the second basic kind of problem: dimension distributed optimization problems and the formulation is as follows:

$$\min F(\vec{x}_i, D), X \in \Omega, i = 1, 2, ..., N$$
$$X = \{\vec{x}_1, \vec{x}_2, ..., \vec{x}_i, ..., \vec{x}_N\} \quad (6)$$

Green planes in Fig. 6 mean the division of objective whereas spaces of data and dimension are the whole. This is the third basic kind of problems: objective distributed optimization problems and the formulation is as follows:

$$\min f_i(X, D), X \in \Omega, i = 1, 2, ..., O$$
$$F = \{f_1, f_2, ..., f_i, ..., f_O\} \quad (7)$$

where $M$ is the number of parts for environmental data, $N$ is the number of parts for variable dimensionalities and $O$ is the number of pasts for objectives. From these three aspects, there are two principles that should be clarified.

**Principle** 1) *Distributed information owned by different agents can be homogeneous or heterogeneous.*

**Principle** 2) *The union sets of distributed information owned by all agents cover the whole space.*

To give an intuitive introduction, categories of distributed optimization problems are given in Fig. 7. Each cube in Fig. 7 represents an agent. Length, width and height represent owned data, dimension and objective respectively. Sub-figure a) demonstrates three agents $a_i$, $a_j$, $a_k$ in data distributed optimization problems. Agents own the same whole dimension and objective spaces whereas data are different. It can be seen that $a_i$ has different data set from both $a_j$ and $a_k$, whereas $a_j$ and $a_k$ have overlapping areas. The similar situation also exits in dimension distributed optimization problems shown in b) and objective distributed optimization problems shown in c). This reflects two principles mentioned above, in which distributed information owned by different agents can be homogeneous (the overlapping areas) or heterogeneous (the separable areas), and the union sets of distributed information owned by all agents cover the whole space. When optimize these three basic optimization problems, it is important to design proper cooperative techniques for distributed and local information learning.

The fourth kind of distributed optimization problems is a mix of three kinds of information. As the name suggests, this kind of distributed optimization problems may have two or even three ($a_r$ in Fig. 7 d)) kinds of distributed information. As shown in sub-figure d), $a_i$, $a_j$, $a_k$ are two-mixed-information distributed optimization problems, which can be formulated as:

$$\min f_i(X, \vec{d}_i), X \in \Omega, i = 1, 2, ..., I$$
$$F = \{f_1, f_2, ..., f_i, ..., f_I\} \quad (8)$$
$$D = \{\vec{d}_1, \vec{d}_2, ..., \vec{d}_i, ..., \vec{d}_I\}$$

$$\min f_j(\vec{x}_j, D), X \in \Omega, j = 1, 2, ..., J$$
$$F = \{f_1, f_2, ..., f_j, ..., f_J\} \quad (9)$$
$$X = \{\vec{x}_1, \vec{x}_2, ..., \vec{x}_j, ..., \vec{x}_J\}$$

$$\min F(\vec{x}_k, \vec{d}_k), X \in \Omega, k = 1, 2, ..., K$$
$$X = \{\vec{x}_1, \vec{x}_2, ..., \vec{x}_k, ..., \vec{x}_K\} \quad (10)$$
$$D = \{\vec{d}_1, \vec{d}_2, ..., \vec{d}_k, ..., \vec{d}_K\}$$

$a_r$ in sub-figure d) is the three-mixed-information distrusted optimization problems and can be formulated as:

$$\min f_r(\vec{x}_r, \vec{d}_r), X \in \Omega, r = 1, 2, ..., R$$
$$F = \{f_1, f_2, ..., f_r, ..., f_R\} \quad (11)$$
$$X = \{\vec{x}_1, \vec{x}_2, ..., \vec{x}_r, ..., \vec{x}_R\}$$
$$D = \{\vec{d}_1, \vec{d}_2, ..., \vec{d}_r, ..., \vec{d}_R\}$$

where $i$, $j$, $k$, $r$ indicate the $i$th, $j$th, $k$th, $r$th agent and $I$, $J$, $K$, $R$ are the total number of agent in different situations. The mixed-information distributed optimization problems are more difficult to optimize since agents should conduct cooperative learning in more information.

*B. DEC for Data Distributed Optimization Problems*

In data distributed optimization problems, there are two situations that get much attention, heterogeneous data

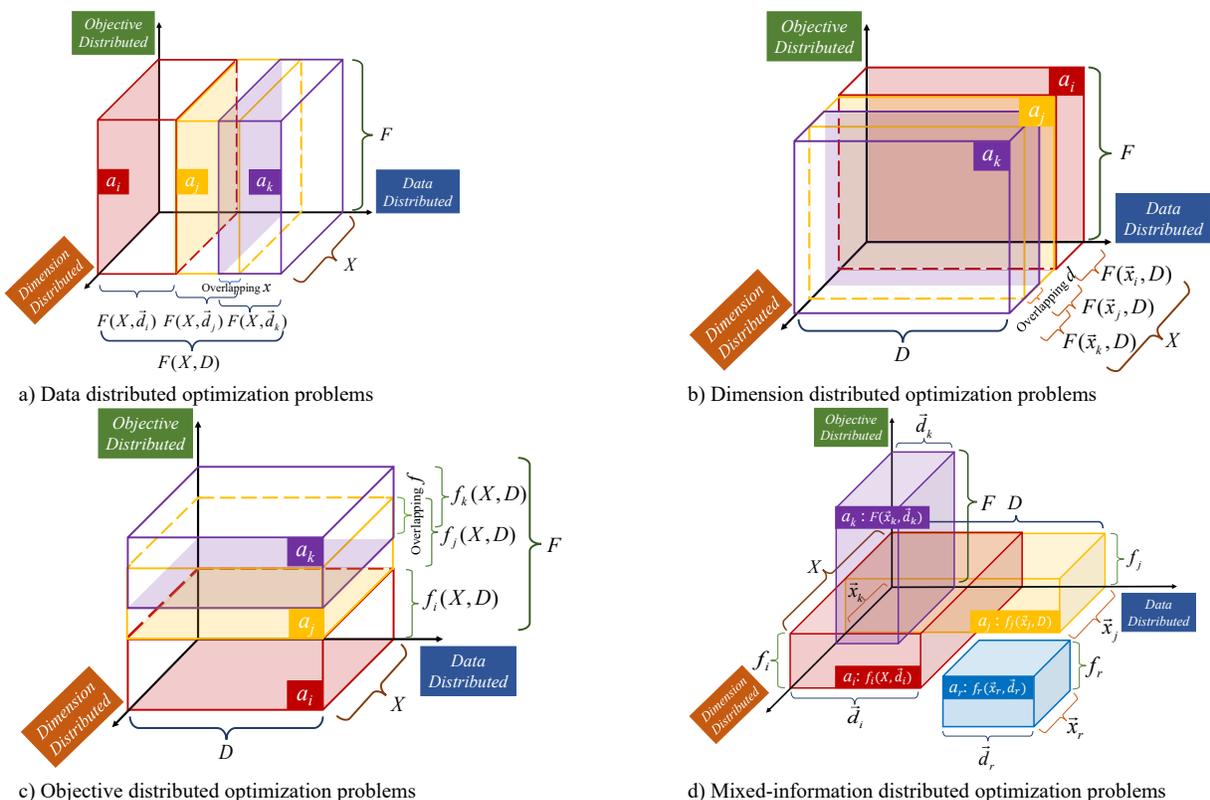

Fig. 7. The taxonomy of distributed optimization problems. a) Data distributed optimization problems. b) Dimension distributed optimization problems. c) Objective distributed optimization problems. d) Mixed-information distributed optimization problems. Each cube represents the information an agent owns, where length, width and height indicate data, dimension and objective.

distributed optimization and privacy data distributed optimization.

*1) DEC for Heterogeneous Data Distributed Optimization*

Since data collected from multiple sources are generally big, dynamic and heterogeneous, they are challenging to handle. Some mining tasks such as $K$-means clustering arise due to distributed data [138], which leads to development of evolutionary cluster techniques under distributed scenarios. Ishibuchi *et al.* [139] proposed a distributed genetics-based machine learning (GBML) approach. It applies EC algorithm to knowledge extraction and data mining, which is used as a fuzzy rule-based classifier designer. The fuzzy GBML can well handle distributed heterogeneous data, which have different missing attributes. Oliveira *et al.* [140] studied scalable datasets and presented two evolutionary scalable metaheuristics in MapReduce, which is named as SF-EAC. The first one aims to enhance $K$-means clusters through evolutionary techniques to handle distributed data. The second one aims to combine clusters into an ensemble. Hajeer *et al.* [141] developed a distributed GA application framework, DEGA-Gen to detect communities in networks. They described in detail how various networks are recorded as chromosomes. Based on DEGA-Gen, Hajeer and Dasgupta [142] designed a distributed encoding technique for GAs to efficiently process distributed data. They tested the proposed methods by distributed datasets generated via Lehigh University Benchmark (LUBM) on Hadoop. Results show that the proposed data-aware Hadoop and evolutionary clustering technique are good at handling heterogeneous data with different structures.

*2) DEC for Privacy Data Distributed Optimization*

Big and distributed data lead to concerns of privacy and security. Federated learning is a distributed data-driven machine learning paradigm, which can reduce privacy and security risks [134]. It cooperates multiple devices to train a global model indirectly based on their local data rather than collects all local data to a central device. Xu *et al.* [131] took the advantage of federated learning and proposed a federated data-driven evolutionary optimization framework for privacy data distributed optimization. On the basis of federated learning, they developed a surrogate-assisted federated data-driven evolutionary algorithm, FDD-EA. All local devices train a radial basis function network and send parameters to the server. The server uses a sorted averaging method to aggregate a global surrogate model to assist evolution. By this way, raw local data are kept secret in local devices, which can prevent data leakage in the process of sending data. Through extensive experiments on distributed environment with noise and non-independently identically distributed (non-IID) data, the proposed method has competitive performance compared with state-of-the-art centralized data-driven surrogate-assisted EAs. Further, Xu *et al.* [143] extended the idea to multi/many-objective optimization and presented a federated data-driven multi/many-objective evolutionary optimization algorithm (FDD-MOEA). This is the first work to combine federated learning with MOEA to preserve data privacy.

*C. DEC for Dimension Distributed Optimization Problems*

There are two situations in dimension distributed optimization problems, nonoverlapping dimension distributed optimization problems and overlapping dimension distributed optimization problems. The latter one is more complex to optimize than the former one.

*1）DEC for Nonoverlapping Dimension Distributed Optimization*

Nonoverlapping is a basic and simple situation in dimension distributed optimization problems. Cooperative coevolution EC algorithms (CCEAs) is a kind of proper EC algorithm for nonoverlapping dimension distributed optimization. However, CCEAs have the decomposition procedure, which is a research hotspot in CCEAs to divide variables into several nonoverlapping subcomponents. On the contrary, dimension distributed optimization problems have the natural characteristic of dimension subcomponents. Therefore, only co-optimization and co-adaption are needed in dimension distributed optimization.

Utkarsh *et al*. [144] studied renewable energy sources (RESs) accompanied with battery storage systems (BSSs) in power distribution and transmission network. They proposed a consensus-based dimension-distributed PSO to optimize ancillary services to transmission system operator (TSO), the technical and economic benefits to distributed network operator (DNO) and RES owners. They implemented the proposed algorithm in a fully distributed manner, in which each node represents an agent and can only communicate with its neighbors. Coadaptation is to use its neighbors current state for fitness evaluation, which is conducted in a synchronization manner among all nodes. Experiments on 30 nodes demonstrate that dimension distributed PSO can real-timely fulfill ancillary service requests of TSO and provide benefits to DNO and RES owners. Besides, Utkarsh *et al*. [133] focused on real-time large-scale optimization problems in smart grids. They further developed a consensus-based dimension distributed PSO algorithm to assist real-time optimal control of renewable-based distribute generators (DGs) and controllable load (CLs) in smart distributed grids. Similarly, coadaptation is to use the current state of neighbors for fitness evaluation in a synchronization manner. The proposed method has better convergence, adaptability and optimality shown through simulations on 30-node and 119-node distribution test systems. Further, they proposed a new gradient module and extended consensus-based dimension distributed PSO algorithm to optimize multi-microgrid system [145].

*2）DEC for Overlapping Dimension Distributed Optimization*

Factored evolutionary algorithms (FEAs) is a recently novel EC paradigm, which factor the original problem into several dimension subsets to optimize [146]. Similar with CCEAs, FEAs have factorization procedure to divide dimensionalities into several subsets. Since dimension distributed optimization problems have the natural characteristic of dimension subcomponents, only competition strategy and sharing strategy are needed when solving this kind of problem. Different from CCEAs, FEAs encourages dimension subsets to overlap with one another, which allow the competition or coevolution with other subsets [147]. Therefore, it is a proper EC paradigm for overlapping dimension distributed optimization.

Fortier *et al*. [148] proposed distributed overlapping swarm intelligence (DOSI) to optimize parameters in artificial neural networks. They applied a swarm to each neuron. Swarms in associated neurons have overlapping parameters to be optimized and can communicate with each other periodically. Swarm with higher fitness can share its information to swarm with lower fitness, while also competing with each other. The lower computational complexity makes DOSI possible to be used in large neural networks. However, distributed manner makes it lack of global perspective. Based DOSI, Fortier *et al*. [147] proposed several multi-swarm algorithms to conduct both full and partial abductive inference in Bayesian belief networks. Each node has a swarm, which can learn state assignment from its Markov blanket. Experiments indicate that DOSI can outperform other non-swarm intelligence methods.

*D. DEC for Objective Distributed Optimization Problems*

Intrinsically, objective-distributed optimization problem is a special kind of multi/many-objective optimization problems, in which objectives are evaluated through different software simulations, authorized organizations or even physical experiments.

Only few researches have been done on DEC for objective-distributed optimization. Oyama *et al*. [149] studied simultaneous structure design of multiple car models and proposed a simultaneous car structure design evolutionary algorithm, which is named as Cheetah. There are three kinds of cars to be simultaneously optimize, MAZDA CX-5 (SUV car), MAZDA6 Wagon (large car), and MAZDA3 Hatchback (small car). Objectives are to minimize the total weight of three kinds of cars and maximize the number of common thickness parts. Software simulations iSPAN and LS-DYNA are used for analyzing obtained designs and crash mode correspondingly, which are distributed from each other. The proposed Cheetah is certificated on K computer [150], which takes more than 360 hours on a supercomputer consisting of more than 80,000 nodes.

Since it takes an unbearable time for hundreds and thousands of fitness evaluations, surrogate models are integrated with EAs, which gives birth of surrogate-assisted EAs (SAEAs) [151]. To maximize the entire wind farm power production and minimize structural loads on individual wind turbines, Yin *et al*. [152] proposed a data-driven multi-objective predictive control approach. Due to intrinsically distributed structure of wind farms and highly time consuming of running computational fluid dynamics (CFD) simulation, they used several models to approximate objectives. The Floris tool is to characterize aerodynamic wake interactions and generate data for predictive control design. It is consisted of three models of wake decay, wake deflection and wake expansion. Wind farm predictor (WFP) is used to predict power and thrust load. It is constructed with support vector machine (SVM) and cooperated with speed-constrained multi-objective PSO (SMPSO) algorithm to generated the optimized yaw angle settings. Experiments show that the proposed method can reduce farm thrust by up to 12.96% while the production can be well maintained. Wei *et al.* [153] proposed a distributed and expensive evolutionary constrained optimization with on-demand evaluation. They optimized distributed problems in which agents have different evaluations for the objective or one constraint. Agents communicate with the server for cooperation since each one only knows partial fitness. The on-demand evaluation strategy saves evaluation budgets for further evolution, which improve performance of the algorithm.

## VII. Challenges and Potential Research Directions

With the development of IoTs and distributed computing paradigms, it becomes common to sense and handle data in a distributed manner. The inherent parallelism of EC provides great potential for DEC to optimize complex problems in such distributed computing environments. Though DEC has been studied for various aspects, it is still facing many challenges and opportunities. The challenges and opportunities give many potential research directions, which can push the development of DEC.

*1) Theoretical Analyses of DEC*

Theoretical analyses like convergence analysis and runtime analysis have long been important and challenging issues for EC. Though some theoretical bases of EC have been studied, in parallel and distributed environment, the characteristics of DEC brings new challenges in theoretical analyses.

i) Convergence Analyses. Convergence is an important aspect of theoretical analysis in EC, which has gain much attention [154], [155]. However, distributed parallelisms propose great challenges for DEC. Firstly, in sub-population parallelism, communication plays a vital role for convergence. High frequency communication enhances global convergence, whereas low frequency communication reduces global convergence. It is significant to study theoretical relationship between communication and convergence in DEC. Further, in sub-population parallelism for CC, the original problem is divided into several sub-problems with different sets of dimensionalities. Optimization results of sub-problems are integrated as the result of the original problem. Therefore, how to make sure sub-problems can converge from the view of theorem is challenging. Besides, in sub-population parallelism for objective coevolution, each population takes charge of one or several objectives. There is no global perspective for all objectives in any population. It needs further studies on how to theoretically guarantee that sub-populations can approach the Pareto front.

ii) Efficiency Analyses. Efficiency is a widely studied topic in EC [44]. Since DEC is originally developed to improve efficiency, some researchers have studied theoretical efficiency analyses in DEC, especially fitness evaluation allocation and communication strategy [66], [26]. Speedup ratio of execution time is an important metric. Generally, execution time $T$ of DEC consists of three parts: $T_1$, evolution time, $T_2$, evaluation time, and $T_3$, communication time. However, distributed computing techniques bring new challenges for efficiency analyses. For example, different computing resources have different execution performance, which may result in different $T_1$ and $T_2$. Besides, communication time $T_3$ is also greatly influenced by transmission. Therefore, providing a fair and robust efficiency analysis metric based on $T$ is worthy to study.

iii) Consistency Analyses. Consistency is an important research point in distributed optimization. For example, dimension distributed optimization problems consist of local objectives and global objectives. Usually, local evaluation and global evaluation cannot achieve consistency. A possible case is that, sub-problems are approaching better objective values with local evaluation, whereas global evaluation of all sub-problems is approaching worse positions. Besides, different sub-problems may have overlapped dimensions. The overlapped dimensions represent interaction of dependent sub-problems. Due to distribution of sub-problems, the optimization of overlapped dimensions may have conflicts in different sub-problems. Thus, how to keep evolution consistency of objectives and dimensions is essential to study.

*2) DEC on New Parallel and Distributed Computing Infrastructures*

With the booming of artificial intelligence (AI), it has been studied to develop artificial intelligence (AI) chips [156] as the research object and Krestinskaya *et al.* [157] have studied automating analogue AI chip design with genetic search. As an important branch of AI, it is promising to study EC and DEC chips since they have wide applications in both academia and industrial engineering. However, how to implement DEC algorithms under limited storage space of chips is a great challenge, especially for large-scale optimization. Fortunately, system-on-chip (SoC) and network-on-chip (NoC) pave the way to develop EC chips. Some researchers have studied to optimize mapping problem in SoC and NoC. For example, Erbas *et al.* [158] investigated multi-objective EAs for the application mapping problem in SoC. Ascia *et al.* [159] addressed topological mapping of intellectual properties by EAs on NoC.

Besides, with the development of parallel and distributed computing techniques, some machine learning and EC platforms have already been proposed. For example, Ribeiro *et al.* [160] constructed ready-to-use Machine Learning as a Service (MLaaS) to provide a scalable, flexible, and non-blocking platform. Tian *et al.* [161] have developed a MATLAB platform for evolutionary multi-objective optimization and Gribble *et al.* [162] have proposed an evolutionary platform for infrastructural services. Inspired by MLaaS and EC platforms, it is a potential to develop DEC platforms as a service on distributed computing infrastructures for large-scale optimization, distributed optimization, mobile optimization, etc. However, when designing DEC platform as a service, there are some problems need to be considered. For example, the relationship between platform processing capacity and task volume, the relationship between storage space of the platform and volume of data to be processed, the relationship between transmission bandwidth and communication frequency and volume, etc.

*3）DEC for Big Data*

EC has been a widely used method for data-driven optimization [151]. With the advent of big data era and development of IoTs, data are distributedly generated and stored, which is hard to fuse in a center. This brings a great challenge for DEC, the isolation of data. Agents only have locally partial data, which is challenging to search for global optima. Xu *et al*. [131], [143] have focused on this challenge and developed federated data-driven EC. Guo *et al*. [163] proposed an edge-cloud co-evolutionary algorithm for distributed data-driven optimization. They train local models based on locally distributed data, which are then fused to aggregate a global model for global optimization. Besides, the heterogeneity of data brings great difficulty as well. It is an ideal case that data are independent and identically distributed (i.i.d.). However, in some cases, data may be non-i.i.d. due to uncertainty of agent behavior and heterogeneity of devices. It is a great challenge to design communication and evaluation strategy among distributed agents. Even worse, some distributed data may be generated by malicious attackers. How to detect the reputation of distributed agents and preprocess distributed data are also worthy to be studied. Apart from the above challenges, it is a potential to mine knowledge from massively distributed data and reuse it, leading to studies of DEC for transfer optimization to solve distributed multitask optimization problems [164].

*4）Privacy and Security of DEC*

With the development and popularity of cloud/edge computing, the deployment of EC on cloud computing platforms can provide a service of robust optimization for various complex problems. In this circumstance, privacy and security attract more attention, leading to another research demand in DEC. Information in DEC such as data, dimensions or objective evaluations are distributed in different agents, which attach importance to privacy protection and are unwilling to leak raw information to others. Therefore, it is significant to study privacy and security of DEC. Some researchers have made attempts to data protection in data distributed optimization. For example, Xu *et al*. [131], [143] combined federated learning with EC to develop federated data-driven EC for privacy protection. Zhao *et al*. [165] proposed a privacy-preserving multi-agent PSO algorithm. They can achieve satisfactory global optimization without data disclosing. Besides, Du *et al*. [166] established a community-structure evolutionary game for privacy protection in social networks. For the future development of DEC, it is also deserving to study privacy protection in dimension distributed and objective distributed optimization problems.

*5）Applications of DEC*

Development of DEC significantly improves efficiency and scalability of EC, which can support us to apply EC for more complex or large-scale optimization problems in both academia and industry. For example, scientific optimization problems such as protein folding, astronomical interferometry, etc. are complex large-scale optimization problems with thousands or even millions of variables. Besides, some industrial applications, such as car model structure optimization, manufacturing optimization, transportation optimization, etc. propose great challenges like expensive evaluation for EC. Fortunately, with the development of DEC and computing techniques, the efficiency and scalability of DEC are greatly improved. It has become a promising method to solve these kinds of problems.

Apart from large-scale and computationally expensive optimization, it is also promising to deploy DEC on distributed optimization applications. For example, in unmanned systems and multi-agent systems (MASs), it is not only a need to use EC algorithms to optimize path planning or scheduling problems but also a great demand to apply DEC to improve autonomy and intelligence of the system, enabling entities to conduct evolutionary operation to cooperatively optimize a mission [167], [168]. However, it is still a big and most concerned challenge to study the consensus theory when apply DEC on unmanned systems and MASs.

VIII. CONCLUSION

With the rapid development of distributed computing paradigms, implementation and realization of EC on distributed computing platforms have attracted increasing attention. This paper has provided a new taxonomy for DEC in terms of purpose, parallel structure of the algorithm, parallel model for implementation and the implementation environment. Specifically, we realize two major purposes of DEC, i.e., improving efficiency through parallel processing and performing distributed optimization by cooperating distributed sub-populations with partial information. Through a comprehensive review, this study shows that DEC can not only take advantages of the natural parallelism of EC to improve search efficiency, but also enable distributed coevolution of spatially-distributed individuals/sub-populations with local variables, data, and objective to achieve scalable distributed optimization. In addition, since distributed optimization is an emerging and attractive trend for DEC, we further give a systematic and formal definition of this kind of problems. The existing studies on DEC for dimension distributed-, data distributed- and objective distributed-optimization problems are reviewed, and their major challenges for DEC are discussed.

The development of Internet of Things and distributed computing platforms like supercomputing, cloud computing and edge computing has made data sensing, processing, and computing ubiquitous. Thanks to the inherent parallelism of EC, it is believed that DEC can become an important technology to solve large-scale and distributed optimization problems in such distributed computing environments, but many challenges and open research questions are still faced. We have discussed some potential research directions on DEC, such as theoretical analyses of DEC, DEC on different distributed computing infrastructures, DEC for big data, privacy and security issues of DEC, and applications of DEC on various large-scale computational-expensive optimization problems like model structure optimization, manufacturing optimization, etc., and distributed optimization problems like unmanned systems, MASs, etc.

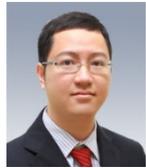

**Wei-Neng Chen** (S'07–M'12–SM'17) received the bachelor's and Ph.D. degrees in computer science from Sun Yat-sen University, Guangzhou, China, in 2006 and 2012, respectively. Since 2016, he has been a Full Professor with the School of Computer Science and Engineering, South China University of Technology, Guangzhou. He has co-authored over 100 international journal and conference papers, including more than 50 papers published in the IEEE Transactions journals. His current research interests include computational intelligence, swarm intelligence, network science, and their applications.

Dr. Chen was a recipient of the IEEE Computational Intelligence Society (CIS) Outstanding Dissertation Award in 2016, and the National Science Fund for Excellent Young Scholars in 2016. He is currently the Vice-Chair of the IEEE Guangzhou Section. He is also a Committee Member of the IEEE CIS Emerging Topics Task Force. He serves as an Associate Editor for the IEEE Transactions on Neural Networks and Learning Systems, and the Complex & Intelligent Systems.

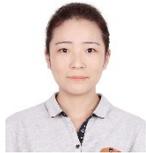

**Feng-Feng Wei** received her Bachelor's degree in computer science from South China University of Technology, Guangzhou, China, in 2019, where she is currently pursuing the Ph.D degree. Her current research interests include evolutionary computation algorithms, multi/many-objective optimization, constrained optimization and their applications on expensive and distributed optimization in real-world problems.

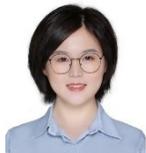

**Tian-Fang Zhao** (S'17-M'21) received the Ph.D. degree in computer science from South China University of Technology, Guangzhou, China, in 2021, and the M.S. degree in computer science from Dalian University of Technology, Dalian, China, in 2017.

She is currently a research fellow in Jinan University, Guangzhou, China. Her current research interests include social media and data mining, complex network, swarm intelligence.

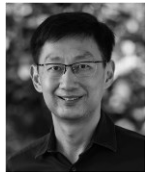

**Kay Chen Tan** (Fellow, IEEE) received the B.Eng. (First Class Hons.) degree in electronics and electrical engineering and the Ph.D. degree from the University of Glasgow, UK, in 1994 and 1997, respectively.

He is currently a Chair Professor of the Department of Computing at The Hong Kong Polytechnic University, Hong Kong, China. He has published over 200 refereed articles and six books, and holds one US patent on surface defect detection.

Prof. Tan is currently the Vice-President (Publications) of IEEE Computational Intelligence Society, USA. He has served as the Editor-in-Chief of IEEE Transactions on Evolutionary Computation from 2015–2020 and IEEE Computational Intelligence Magazine from 2010–2013, and currently serves as the Editorial Board Member of over 10 journals. He is currently an IEEE Distinguished Lecturer Program (DLP) Speaker and Chief Co-Editor of Springer Book Series on Machine Learning: Foundations, Methodologies, and Applications.

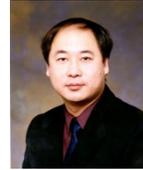

**Jun Zhang** (M'02–SM'08–F'17) received the Ph.D. degree in Electrical Engineering from the City University of Hong Kong in 2002. Currently, he is a visiting professor of Division of Electrical Engineering, Hanyang University. His research interests include computational intelligence, cloud computing, data mining, and power electronic circuits.

He has published over 200 technical papers in his research area. Dr. Zhang was a recipient of the China National Funds for Distinguished Young Scientists from the National Natural Science Foundation of China in 2011 and the First-Grade Award in Natural Science Research from the Ministry of Education, China, in 2009. He is currently an Associate Editor of the IEEE Trans. on Evolutionary Computation and IEEE Trans. on Cybernetics.